\documentclass[conference]{IEEEtran}  
\IEEEoverridecommandlockouts                              

\usepackage{cite}
\usepackage{graphicx} 
\usepackage{textcomp}
\usepackage{algorithmic}
\usepackage{epsfig} 
\usepackage{amsmath} 
\usepackage{amssymb, amsfonts}  
\usepackage{booktabs}
\usepackage{siunitx}
\usepackage{soul,framed}
\usepackage{color}
\usepackage[table]{xcolor}
\def\BibTeX{{\rm B\kern-.05em{\sc i\kern-.025em b}\kern-.08em
    T\kern-.1667em\lower.7ex\hbox{E}\kern-.125emX}}

\usepackage{arydshln}
\usepackage[flushleft]{threeparttable}

\definecolor{darkgreen}{rgb}{0,0.6,0.2}

\begin{document}

\title{
Dense Dilated Convolutions Merging Network for Semantic Mapping of Remote Sensing Images
}

\makeatletter
\newcommand{\linebreakand}{%
  \end{@IEEEauthorhalign}
  \hfill\mbox{}\par
  \mbox{}\hfill\begin{@IEEEauthorhalign}
}
\makeatother

\author{\IEEEauthorblockN{1\textsuperscript{st} Qinghui liu}
\IEEEauthorblockA{\textit{SAMBA and Machine Learning Group} \\
\textit{Norwegian Computing Center and UiT}\\
Oslo, Norway \\
Brian.Liu@nr.no}
\and
\IEEEauthorblockN{2\textsuperscript{nd} Michael Kampffmeyer}
\IEEEauthorblockA{\textit{Machine Learning Group} \\
\textit{UiT, the Arctic University of Norway}\\
Troms{\o}, Norway\\
Michael.C.Kampffmeyer@uit.no}
\and
\IEEEauthorblockN{3\textsuperscript{nd} Robert Jenssen}
\IEEEauthorblockA{\textit{Machine Learning Group} \\
\textit{UiT, the Arctic University of Norway}\\
Troms{\o}, Norway \\
Robert.Jenssen@uit.no}
\linebreakand 
\IEEEauthorblockN{4\textsuperscript{th} Arnt-B{\o}rre Salberg}
\IEEEauthorblockA{\textit{dept. SAMBA of NR} \\
\textit{Norwegian Computing Center}\\
Oslo, Norway \\
Arnt-Borre.Salberg@nr.no}

\thanks{This  work  is  supported  by  the  foundation  of  the  Research Council of Norway under Grant 220832 and Grant 239844.}%
}

\maketitle

\IEEEpubid{\begin{minipage}{\textwidth}\ \\[12pt] \centering
~�\\~�\\~\\  
978-1-7281-0009-8/19/\$31.00 \copyright 2019 IEEE  
\end{minipage}}

\begin{abstract}
We propose a network for semantic mapping called the Dense Dilated Convolutions Merging Network (DDCM-Net) to provide a deep learning approach that can recognize multi-scale and complex shaped objects with similar color and textures, such as buildings, surfaces/roads, and trees in very high resolution remote sensing images. The proposed DDCM-Net consists of dense dilated convolutions merged with varying dilation rates. This can effectively enlarge the kernels' receptive fields, and, more importantly, obtain fused local and global context information to promote surrounding discriminative capability.
We demonstrate the effectiveness of the proposed DDCM-Net on the publicly available ISPRS Potsdam dataset and achieve a performance of 92.3\% F1-score and 86.0\% mean intersection over union accuracy by only using the RGB bands, without any post-processing. We also show results on the ISPRS Vaihingen dataset, where the DDCM-Net trained with IRRG bands, also obtained better mapping accuracy (89.8\% F1-score) than previous state-of-the-art approaches.

\end{abstract}


\section{Introduction}
Automatic semantic interpretation of remote sensing images is important for a wide range of practical applications, such as urban land cover classification \cite{kampffmeyer2016semantic}, traffic monitoring and vehicle detection.
Large-scale semantic mapping is a challenging task which consists of the assignment of a semantic category to every pixel in very high resolution (VHR) aerial images. Due to the successes of deep learning methods, a large variety of modern approaches to pixel-to-pixel classification are based on deep convolutional neural networks (CNN), in particular end-to-end learning with fully convolutional neural networks (FCN) \cite{long2015fully}. However, to achieve higher performance, CNN and FCN based methods \cite{zhao2016pyramid, wang2017understanding, peng2017large} normally rely on deep multi-scale architectures which typically require a large number of trainable parameters and computation resources. 

 \begin{figure}[tbp]
 \centering
  \includegraphics[width=0.47\textwidth]{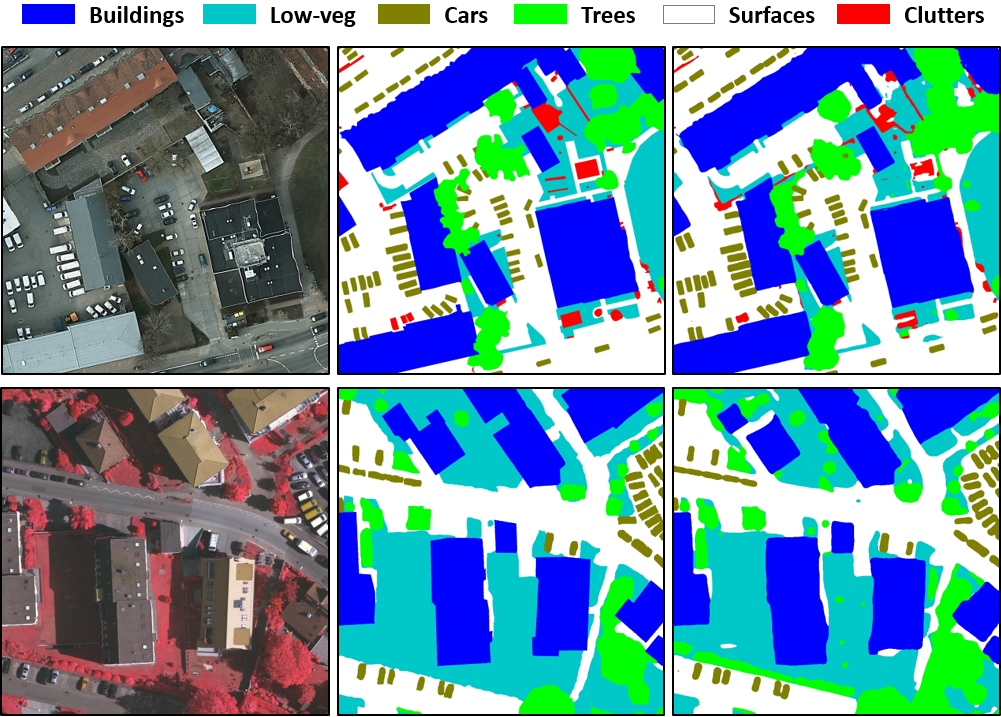} 
  \caption{Examples of semantic mapping of remote sensing images on RGB (top-left) and IRRG data (bottom-left) respectively with our DDCM network. From left to right, test images, ground truths, and mapping results.}
  \label{fig:semapping}
\end{figure}

In this work, we propose a novel network architecture, called the dense dilated convolutions merging network (DDCM-Net), which utilizes multiple dilated convolutions merged with various dilation rates.
The proposed network learns with densely linked dilated convolutions and outputs a fusion of all intermediate features without losing resolutions during the extraction of multi-scale features. Our experiments demonstrate that the network achieves robust and accurate results with relatively few parameters and layers. Fig. \ref{fig:semapping} shows illustrative examples of semantic mapping results on RGB and IRRG data respectively with our DDCM-Net methods. These results will be further discussed in section \ref{exp}.



\section{Methods}
We first briefly revisit dilated convolutions which are used in DDCM networks. We then present our proposed DDCM architecture and provide training details.

\subsection{Dilated Convolutions}
Dilated convolutions~\cite{yu2015multi} have been demonstrated to improve performance in many classification and segmentation tasks \cite{chen2018deeplab, pelt2018mixed, li2018csrnet, wei2018revisiting}. 
One key advantage is that they allow us to flexibly adjust the filter's receptive field to capture multi-scale information without resorting to down-scaling and up-scaling operations. A 2-D dilated convolution operator can be defined as

\begin{equation}\label{dconv} 
   g_{i,j}(x_{\ell}) = \sum_{c=0}^{C_{\ell}} \theta^{i,j}_{k, r} \ast x_{\ell}^{c}
\end{equation}
where, $\ast$ denotes a convolution operator, $g_{i,j}:\mathbb{R}^{H_{\ell}\times W_{\ell}\times C_{\ell}} \rightarrow\mathbb{R}^{H_{\ell+1}\times W_{\ell+l}}$ convolves the input feature map $x_{\ell}\in \mathbb{R}^{H_{\ell}\times W_{\ell}\times  C_{\ell}}$ within channel $c \in \{0,1,\dotsc, C_{\ell}\}$ at row $i$ and column $j$. A dilated convolution $\theta_{k, r}$ with a filter $k$ and dilation $r \in \mathbb{Z}^+$ is only nonzero for a multiple of $r$ pixels from the center. In dilated convolution, a kernel size $k$ is enlarged to $k+(k-1)(r-1)$ with the dilation factor $r$. As a special case, a dilated convolution with dilation rate $r=1$ corresponds to a standard convolution. 



\subsection{Dense Dilated Convolutions Merging Module}
The proposed dense dilated convolutions merging (DDCM) module densely stacks multi-scale features and merges them to yield more accurate and robust representations with fewer parameters. Fig. \ref{fig:ddcm} illustrates the basic structure of the DDCM module. 

DDCM module consists of a number of Dilated CNN-stack (DCs) blocks with a merging module as output. A basic DCs block is composed of a dilated convolution followed by PReLU \cite{he2015delving} non-linear activation and batch normalization (BN) \cite{ioffe2015batch}. It then stacks the output with its input together to feed the next layer, which can alleviate context information loss and problems with vanishing gradients when adding more layers. The final network output is computed by a merging layer composed of $1\times1$ filters with BN and PReLU in order to efficiently combine all stacked features generated by intermediate DCs blocks. 

In a DDCM module, all feature maps are maximally utilized with high computational efficiency while preserving the input resolution throughout the network. In particular, densely connected DCs blocks, typically configured with linearly increasing dilation factors, enable DDCM networks to have very large receptive fields with just a few layers as well as to capture rich global representations by merging multi-scale features properly.

 \begin{figure}[htbp]
 \centering
  \includegraphics[width=0.48\textwidth]{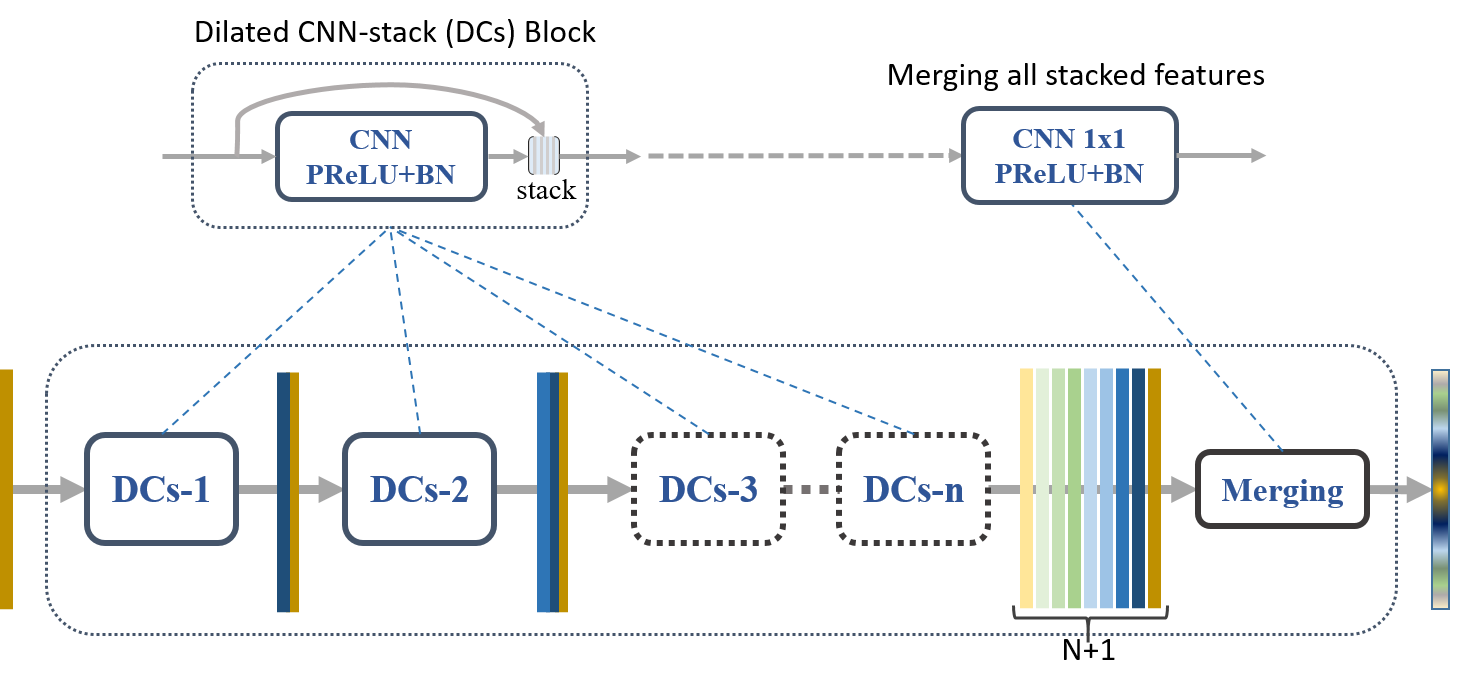} 
  \caption{Example of the DDCM architecture composed of $n$ DC blocks with various dilation rates $\{1, 2, 3, ... , n\}$.}
  \label{fig:ddcm}
\end{figure}

 \begin{figure}[htbp]
 \centering
  \includegraphics[width=0.48\textwidth]{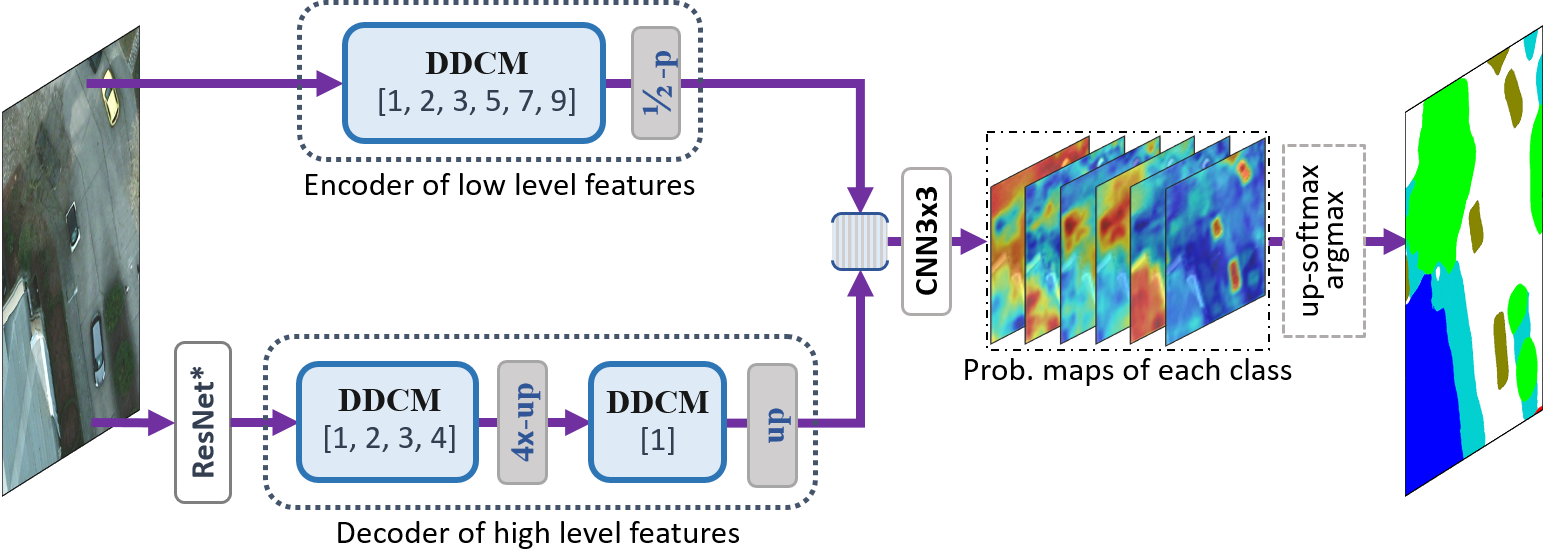} 
  \caption{End-to.end pipeline of DDCM-Net for semantic mapping of VHR images. The encoder of low level features encodes multi-scale contextual information from the initial input images by a DDCM module (output 3-channel) using $3\times3$ kernels with 6 different dilation rates $[1, 2, 3, 5, 7, 9]$. The decoder of high level features decodes highly abstract representations learned from ResNet50 (output 1024-channel) by 2 DDCM modules with rates $[1, 2, 3, 4]$ (output 36-channel) and $[1]$ (output 18-channel) separately. The transformed low-level and high-level feature maps by DDCMs are then fused together to infer pixel-wise class probabilities. Here, 'p' and 'up' denote pooling and up-sampling respectively.}
  \label{fig:ddcm_resnet}
\end{figure}


Fig. \ref{fig:ddcm_resnet} shows the end-to-end pipeline of DDCM network (DDCM-Net) architecture combined with a pre-trained ResNet \cite{he2016deep} for semantic mapping tasks. The proposed DDCM-Net is easy to implement, train and combine with existing architectures. In our work, we only utilize the first 3 bottleneck layers of ResNet50 and remove the last bottleneck layer and fully connected layers to reduce the number of parameters to train. 





\subsection{Data augmentation and normalization}
We randomly sample 5000 image patches of size $256\times 256$ in run time from VHR training images (of size $6000\times 6000$) for each training epoch and flip and mirror images for data augmentation. These patches are normalized to [0.0, 1.0] by dividing by 255 for all bands (RGB, or IRRG). We used a pre-trained ResNet50 model that has been trained on ImageNet \cite{ILSVRC15}. No mean and standard deviation normalization were used. 




\subsection{Optimizer and weighted loss function}
In our work, we choose Adam \cite{KingmaB14adam} with AMSGrad \cite{amsgrad2018} as the optimizers with weight decay $5 \times 10^5$ and polynomial learning rate (LR) decay $(1 - \frac{cur\_iter}{max \_iter})^{0.9}$ with the maximum iterations of $10^8$ for the model. We also set $2 \times LR$ to all bias parameters in contrast to weights parameters. Guided by our empirical results, we use an initial LR of $\frac{8.5 \times 10^{-5}}{\sqrt{2}} $ and step-LR schedule method which drops the LR by 0.85 at every 15 epochs. As loss function, we apply a cross-entropy loss function with median frequency balancing as defined in \cite{kampffmeyer2016semantic}. 




%
%
%

\section{Experiments}
\label{exp}
We investigate the proposed network on the ISPRS 2D semantic labeling dataset \cite{ISPRS2018} which is comprised of very high resolution aerial images over two cities: Potsdam and Vaihingen in Germany. In this work, we only use RGB bands of Potsdam dataset and IRRG (Infrared-Red-Green) bands of Vaihingen dataset.

\subsection{Dataset}
The ISPRS Potsdam dataset contains 38 RGB images ($6000\times 6000$) annotated with six different labels including impervious surfaces, buildings, trees, low vegetation, cars and clutters. Originally, 24 of the images were public available and 14 were included in a hold-out test set. The Vaihingen dataset has 33 IRRG images, where 16 were from the original public dataset, and 17 were included the hold-out test set. To evaluate our models, the original public part (24 images) of the Potsdam dataset was divided into training, validation (areas: 4\_10 and 7\_10), and local test set (areas: 5\_11, 6\_9 and 7\_11). The the original public part (16 images) of the Vaihingen dataset was similarly split into training, validation (tiles of 7 and 28), and local test set (tiles of 5, 15, 21, and 30). Please note that our trained models are evaluated both on the local test sets to compare with our previous work \cite{kampffmeyer2016semantic, liuqinghui2018}, and on the hold-out test sets to compare with other related published work.

\subsection{Evaluation methods}
We train and validate all networks with patches of size $256\times 256$ as input and batch size of 5. All hyper-parameters settings, except the learning rates, were shared for the different models. At test time, we apply test time augmentation (TTA) in terms of flipping and mirroring. We use sliding windows (with $448 \times 448$ size at a 100px stride) on a test image and stitch the results together by averaging the predictions of the over-lapping TTA regions to form the output. The performance is measured by both the F1-score \cite{kampffmeyer2016semantic}, and mean Intersection over Union  (IoU) \cite{liuqinghui2018}. 

\subsection{Results}
Table \ref{tab:rgbresults} shows our results on the hold-out test sets and our local test sets of ISPRS Potsdam and Vaihingen separately with a single trained model. The mean F1-score (mF1) and the mean IoU (mIou) are computed as the average measure of all classes except the clutter class. Fig. \ref{fig:test} shows a qualitative comparison of the semantic mapping results from our model and the ground truths.  

\begin{table}[hptb!]
\centering 
  \caption{Results on the hold-out test images of ISPRS Potsdam and Vaihingen datasets with a single trained DDCM-R50 model separately.}
\resizebox{\columnwidth}{!}{
\begin{threeparttable}
\begin{tabular}{c|p{8mm}|p{8mm}p{8mm}p{11mm}p{9mm}p{9mm}|p{9mm}} 
    \textbf{Potsdam} & $\textbf{Avg.}$ & \textbf{Building} & \textbf{Tree} & \textbf{Low-veg} & \textbf{Surface} & \textbf{Car}  & \textbf{OA} \\  \hline  
    F1-score & 0.923 \newline 0.925$^{*}$ & 0.969 \newline 0.983$^{*}$ & 0.894 \newline 0.892$^{*}$ & 0.877 \newline 0.865$^{*}$ & 0.929 \newline 0.946$^{*}$ & 0.949 \newline 0.939$^{*}$ & 0.908 \newline0.931$^{*}$\\ \hdashline
    IoU & 0.860 \newline 0.863$^{*}$ & 0.940 \newline 0.966$^{*}$ & 0.809 \newline 0.805$^{*}$ & 0.781 \newline 0.762$^{*}$ & 0.867 \newline 0.898$^{*}$ & 0.902 \newline 0.885$^{*}$&  \\  \bottomrule
    \textbf{Vaihingen} & $\textbf{}$ & \textbf{} & \textbf{} & \textbf{} & \textbf{} & \textbf{}  & {} \\  \hline  
    F1-score & 0.898 \newline 0.909$^{*}$ & 0.953 \newline 0.973$^{*}$ & 0.894 \newline 0.914$^{*}$ & 0.833 \newline 0.814$^{*}$ & 0.927 \newline 0.934$^{*}$ & 0.883 \newline 0.909$^{*}$ & 0.904 \newline 0.921$^{*}$ \\ \hdashline
    IoU & 0.817 \newline 0.837$^{*}$ & 0.909 \newline 0.948$^{*}$ & 0.808 \newline 0.842$^{*}$ & 0.713 \newline 0.686$^{*}$ & 0.863 \newline 0.876$^{*}$ & 0.790 \newline 0.832$^{*}$ &  \\  \hline
\end{tabular}
\begin{tablenotes}
            \item[*] Results were measured on our local test images.
    \end{tablenotes}
  \end{threeparttable}
} 
\label{tab:rgbresults}%

\end{table}

\begin{figure}[htpb!]
 \centering
  \includegraphics[width=0.48\textwidth]{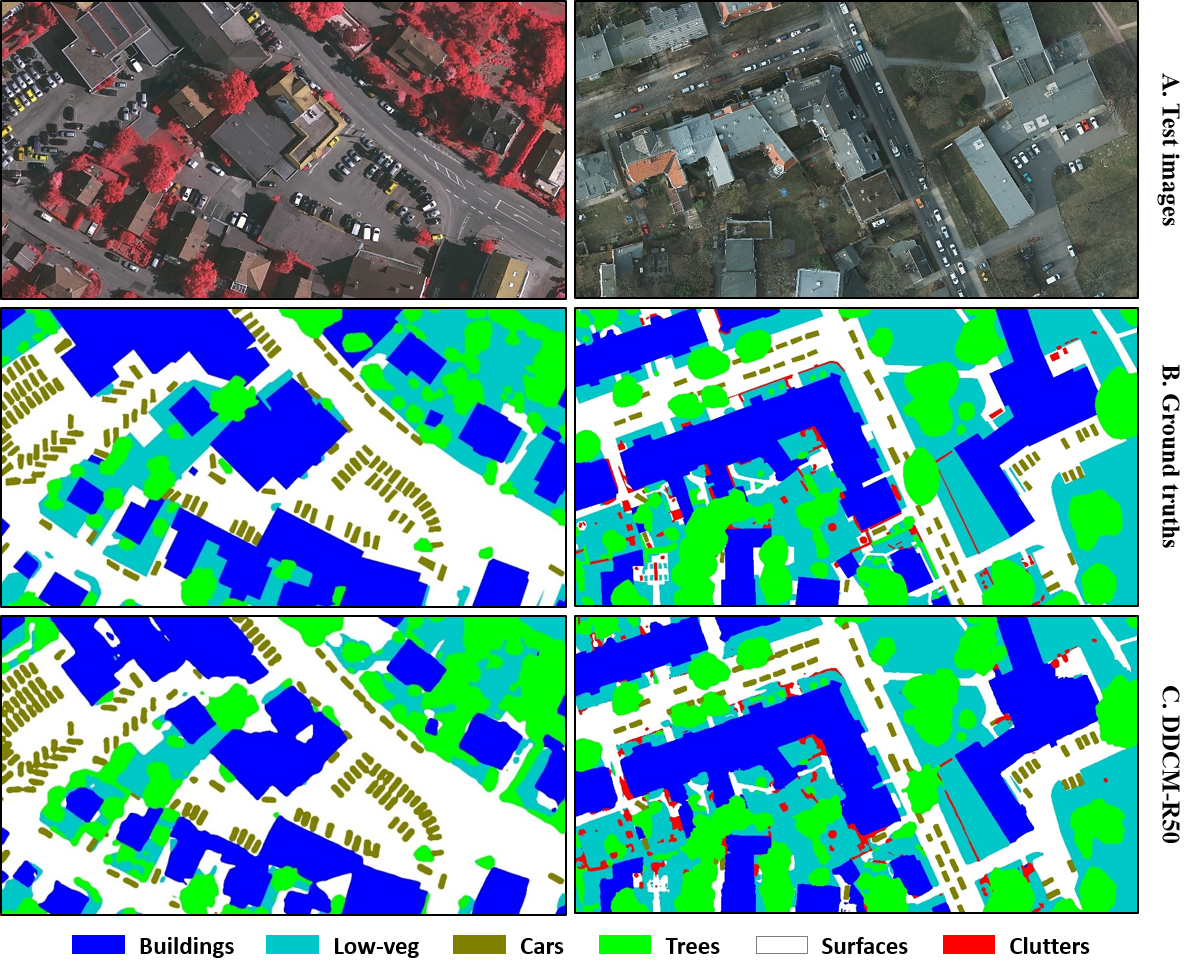} 
  \caption{Mapping results for test images of Vaihingen tile-27 (left) and Potsdam tile-3\_14 (right). A. Test images (top), B. the Ground truths (center), C. DDCM-R50 (bottom).}
  \label{fig:test}
\end{figure}

We also compare our results to other related published work on the ISPRS Potsdam RGB dataset and Vaihingen IRRG dataset. These results are shown in Table \ref{tab:testresults2} and \ref{tab:testresults4} respectively. Our single model with overall F1-score (92.3\%) on Potsdam RGB dataset, achieves around 0.5 percent higher than the secondary best model - FuseNet+OSM \cite{audebert2017joint} which used OpenStreetMap (OSM) as an additional data source. In other words, our model achieves better performance with fewer labeled training data. Similarly, our model trained on Vaihingen IRRG images, also obtained the best overall performance with 89.8\% F1-score which is around 1.1\% higher than the second best model GSN \cite{wang2017gated}. It's worth noting that our model is the only one that works equally well on both Vaihingen IRRG dataset and Potsdam RGB dataset, which outperforms the DST\_2 \cite{Sherrah16} model with 3.9\% and 0.6\% higher F1-score on Vaihingen and Potsdam dataset respectively.

\begin{table}[hptb!]
\centering 
  \caption{Comparisons between our method with other published methods on the hold-out RGB test images of ISPRS Potsdam dataset.}
\resizebox{\columnwidth}{!}{
\begin{threeparttable}
\begin{tabular}{c|p{9mm}|p{8mm}p{8mm}p{11mm}p{9mm}p{9mm}|p{8mm}} \hline
    \textbf{Models} & $\textbf{OA}$ & \textbf{Building} & \textbf{Tree} & \textbf{Low-veg} & \textbf{Surface} & \textbf{Car} & \textbf{mF1} \\   \hline
    HED+SEG.H-Sc1 \cite{MarmanisSWGDS16} & 0.851  & 0.967  & 0.686  & 0.842 & 0.850  & 0.858  & 0.846 \\  
    RiFCN \cite{MouRiFCN2018} & 0.883  & 0.930  & 0.819  & 0.837 & 0.917  & 0.937  & 0.861 \\
    RGB+I-ensemble \cite{michael2018} & 0.900  & 0.936  & 0.845 & 0.822  & 0.870 & 0.892 & 0.873 \\
    Hallucination \cite{michael2018} & 0.901  & 0.938  & 0.848 & 0.821  & 0.873 & 0.882  & 0.872 \\
    SegNet RGB \cite{audebert2017joint} & 0.897  & 0.929  & 0.851  & 0.850 & 0.930  & 0.951  & 0.902\\ 
    DST\_2 \cite{Sherrah16} & 0.903  & 0.964  & 0.880  & 0.867 & 0.925  & 0.947 & 0.917 \\
    FuseNet+OSM \cite{audebert2017joint} & \cellcolor{gray!25} \textbf{0.923}  & 0.959  & 0.851  & 0.863 & \cellcolor{gray!25}\textbf{0.953}  & \cellcolor{gray!25}\textbf{0.968}  & 0.918\\ \hdashline
    DDCM-R50 (ours) & 0.908 \newline (-1.5\%) & \cellcolor{gray!25}\textbf{0.969} \newline (+0.5\%)   &\cellcolor{gray!25}\textbf{0.894} \newline (+1.4\%)  & \cellcolor{gray!25}\textbf{0.877} \newline (+1.0\%)  & 0.929\newline (-2.4\%)  & 0.949\newline (-1.9\%)  & \cellcolor{gray!25}\textbf{0.923} \newline (+0.5\%)\\  \hline
\end{tabular}
  \end{threeparttable}
} 
\label{tab:testresults2}%

\end{table}

\begin{table}[hptb!]
\centering 
  \caption{Comparisons between our method with other published methods on the hold-out IRRG test images of ISPRS Vaihingen Dataset.} 
  
\resizebox{\columnwidth}{!}{
\begin{threeparttable}
\begin{tabular}{c|p{8mm}|p{8mm}p{9mm}p{11mm}p{9mm}p{9mm}|p{8mm}} \hline
    \textbf{Models} & $\textbf{OA}$ & \textbf{Building} & \textbf{Tree} & \textbf{Low-veg} & \textbf{Surface} & \textbf{Car}  & \textbf{mF1} \\   \hline
    UOA \cite{lin2016efficient} & 0.876  & 0.921  & 0.882  & 0.804 & 0.898  & 0.820  & 0.865 \\  
    ADL\_3 \cite{paisitkriangkrai2015effective} & 0.880  & 0.932  & 0.882  & 0.823 & 0.895  & 0.633  & 0.833 \\ 
    DST\_2 \cite{Sherrah16} & 0.891  & 0.937  & 0.892  & 0.834 & 0.905  & 0.726 & 0.859 \\ 
    ONE\_7 \cite{audebert2016semantic} & 0.898  & 0.945  & 0.899  & \cellcolor{gray!25}\textbf{0.844} & 0.910  & 0.778 & 0.875\\ 
    DLR\_9 \cite{MarmanisSWGDS16} & 0.903  & 0.952  & 0.899  & 0.839 & 0.924  & 0.812  & 0.885 \\  
    GSN \cite{wang2017gated} & 0.903  & 0.951  & \cellcolor{gray!25}\textbf{0.899}  & 0.837 & 0.922  & 0.824  & 0.887 \\\hdashline
    DDCM-R50 (ours) & \cellcolor{gray!25}\textbf{0.904} \newline (+0.1\%) & \cellcolor{gray!25}\textbf{0.953} \newline (+0.1\%)   &0.894 \newline (-0.5\%)  & 0.833 \newline (-1.1\%)  & \cellcolor{gray!25}\textbf{0.927}\newline (+0.3\%) & \cellcolor{gray!25}\textbf{0.883}\newline (+5.9\%) & \cellcolor{gray!25}\textbf{0.898} \newline (+1.1\%)\\  \hline
\end{tabular}
  \end{threeparttable}
} 
\label{tab:testresults4}%

\end{table}

In addition, we evaluate our method on the local Potsdam test set to compare with other popular architectures reviewed and re-implemented in \cite{liuqinghui2018}. Our DDCM-R50 model achieved the highest mIoU score (80.8\%) compared to others while using much fewer parameters and computational cost (FLOPs) as shown in Table \ref{tab:parameters}. Note that the performance on full reference ground truths is slightly lower than on eroded boundary ground truths as the boundary pixels are not ignored during evaluation.

\begin{table}[hptb!]
\centering 
  \caption{Quantitative Comparison of parameters size, FLOPs (measured on input image size of $3 \times 256 \times 256$), and mIoU  on ISPRS Potsdam RGB dataset.}
\resizebox{0.95\columnwidth}{!}{
\begin{threeparttable}
\begin{tabular}{c|c|p{18mm}p{18mm}p{18mm}} \hline 
\textbf{Models} & \textbf{Backbones} & \textbf{Parameters} \newline (Million)& \textbf{FLOPs} \newline (Giga) & \textbf{mIoU}$^{*}$  \\  \hline
UNet \cite{ronneberger2015u}& VGG16 & 31.04 & 15.25 & 0.715\\
FCN8s \cite{long2015fully} & VGG16 & 134.30 & 73.46 & 0.728\\
SegNet \cite{badrinarayanan2015segnet}& VGG19 & 39.79 & 60.88 & 0.781\\
GCN \cite{peng2017large} & ResNet50 & 23.84 & 5.61  & 0.774\\
PSPNet \cite{zhao2016pyramid} & ResNet50 & 46.59 & 44.40 & 0.789\\
DUC \cite{wang2017understanding} & ResNet50 & 30.59 & 32.26 & 0.793\\  \hdashline
DDCM-R50 (ours) & ResNet50 & \cellcolor{gray!25}\textbf{9.99}  & \cellcolor{gray!25}\textbf{4.86} &\cellcolor{gray!25} \textbf{0.808}\\ \hline 
\end{tabular}
    \begin{tablenotes}
            \item[*] mIoU was measured on full reference ground truths of our local test images 5\_11, 6\_9 and 7\_11 in order to fairly compare with our previous work \protect\cite{liuqinghui2018}.
    \end{tablenotes}
  \end{threeparttable}
}
\label{tab:parameters}%
\end{table}

\section{Conclusions}
In this paper, we presented a dense dilated convolutions merging (DDCM) network architecture for semantic mapping in very high-resolution aerial images. 
The proposed architecture applies dilated convolutions to learn features at varying dilation rates, and merges the feature map of each layer with the feature maps from all previous layers. 
On both the Potsdam and Vahingen datasets, the DDCM-Net architecture achieves the best mean $F_1$ score compared to the other architectures, but with much fewer parameters and feature maps. 
DDCM-Net is easy to adapt to address a wide range of different problems by using various combinations of dilation rates, is fast to train, and achieves accurate results even on small datasets.








\end{document}